# Multilingual transformer and BERTopic for short text topic modeling: The case of Serbian

Darija Medvecki[1] [0000-0002-4180-0050], Bojana Bašaragin[1] [0000-0002-7679-1676], Adela Ljajić[1] [0000-0001-7326-059X] and Nikola Milošević[1] [0000-0003-2706-9676]

[1] The Institute for Artificial Intelligence Research and Development of Serbia, Fruškogorska 1, 21000 Novi Sad, Serbia
`{darija.medvecki,bojana.basaragin,adela.ljajic,nikola.milosevic}`
`@ivi.ac.rs`

**Abstract.** This paper presents the results of the first application of BERTopic, a state-of-the-art topic modeling technique, to short text written in a morphologically rich language. We applied BERTopic with three multilingual embedding models on two levels of text preprocessing (partial and full) to evaluate its performance on partially preprocessed short text in Serbian. We also compared it to LDA and NMF on fully preprocessed text. The experiments were conducted on a dataset of tweets expressing hesitancy toward COVID-19 vaccination. Our results show that with adequate parameter setting, BERTopic can yield informative topics even when applied to partially preprocessed short text. When the same parameters are applied in both preprocessing scenarios, the performance drop on partially preprocessed text is minimal. Compared to LDA and NMF, judging by the keywords, BERTopic offers more informative topics and gives novel insights when the number of topics is not limited. The findings of this paper can be significant for researchers working with other morphologically rich low-resource languages and short text.

## 1 Introduction

As an unsupervised task, topic modeling is an invaluable tool in many areas, especially where user-generated content (emails, user comments, reviews, complaints, etc.) needs to be analyzed without prior annotation. While there is significant work done in this area for English, especially using Latent Dirichlet Allocation (LDA) [1], a classical topic modeling method, this is an unexplored area for Serbian. The authors in [2] initiated this work by applying LDA and Nonnegative Matrix Factorization (NMF) [3] on tweets in Serbian to find the hidden reasons for COVID-19 vaccine hesitancy.

Recently, BERTopic [4] has been proposed as a new, more flexible model for detecting topics in unannotated text. Unlike more conventional methods that use bag-of-words approaches to describe documents, BERTopic uses state-of-the-art pre-trained language models to create document embeddings, which enables capturing semantic relationships between words. As this framework relies on context, by definition, it should not require substantial data preprocessing. In contrast, both LDA and NMF require extensive preprocessing and significant parameter tuning. Some preliminary research showed that BERTopic generalizes better than LDA judging by topic coherence [5, 6] and topic diversity scores [6], and that it creates more clear-cut topics and gives more novel insights compared to NMF and LDA [7].

Since BERTopic has not been applied to Serbian yet, our aim was to explore its usability and performance, particularly on short minimally preprocessed text. As a morphologically rich language, Serbian normally requires the lemmatization step for most NLP tasks. Using pre-trained language models as embedding models for BERTopic could render this step unnecessary. Our results can serve as pointers for other researchers working with short text and morphologically rich languages.

## 2    Related work

Several methods can provide insight into structures hidden in large amounts of text by grouping them into topics. Two of the most used traditional topic modeling methods are Latent Dirichlet Allocation (LDA) [1] and Nonnegative Matrix Factorization (NMF) [3]. As language-agnostic models, both have been applied in the context of different languages and texts of various lengths. The downside to these models, often mentioned in research papers, can be summarized into three points. First, they have difficulties modeling short text due to the data sparseness problem [8]. Second, LDA and NMF both require the number of topics as one of their initial parameters. Finding this number requires extensive parameter tuning, making the models difficult to optimize [9]. Third, both models require significant preprocessing, including stemming and/or lemmatization, which can produce unreliable and ambiguous results, depending on the language and the quality of the algorithms used [10].

The rise of self-attention-based models and the concept of pre-training in the late 2010s and early 2020s gave rise to a number of pre-trained language models (PLMs). PLMs made significant advances in many fields of natural language processing by introducing pre-trained contextual embeddings. BERTopic [4] is the most recent topic modeling method that leverages PLMs to create document embeddings. Combining such embeddings with a class-based TF-IDF procedure allows BERTopic to better deal with sparse data.

So far, BERTopic has been used in diverse domains (hospitality, sports management, finance, and medicine) to gain insight into different types of text (customer reviews, students' answers, consumer, and general domain data), mostly in English [11–14]. Unlike LDA or NMF, applying BERTopic to other languages requires language-specific sentence embeddings, an issue that can be overcome by leveraging monolingual or multilingual PLMs that fit the specific use case. In a pilot study by [5], BERTopic was tested on news texts using several mono- and multilingual PLMs trained on Arabic data. Authors of [15] applied BERTopic that employs AraBERT PLM to tweets as one of the steps in designing a cognitive distortion classification model.

Thanks to its ability to integrate multilingual PLMs, researchers in [16] used BERTopic to make a hoax news classification pipeline for Indonesian.

When applying BERTopic to news and tweets along with NMF and LDA, authors in [4] reported high topic coherence scores across all datasets, with the highest ones on slightly preprocessed text of tweets. Reseachers in [6] compared the performance of LDA and BERTopic with two clustering algorithms (HDBSCAN and k-means) on student comments and news and found that BERTopic using HDBSCAN achieved the highest topic coherence and topic diversity scores. Authors in [5] found that BERTopic with AraVec2.0 as a word embedding model outperformed NMF, LDA and other BERTopic embedding models in terms of NPMI topic coherence scores.

The results achieved by BERTopic so far suggest that it is a powerful model able to overcome the difficulties of more traditional options.

## 3      Methodology

We started the process of exploring BERTopic by defining the research questions. After applying several architecture variants, we specified the architecture that would best fit our needs and the dataset. Since the first study of using topic modeling on Serbian short text was performed on tweets, we used the same dataset as an opportunity to compare BERTopic against LDA and NMF on the same data.

### 3.1     Research questions

There were two research questions we wanted to address:

**RQ 1: How does preprocessing affect the quality of BERTopic topic representations in case of a morphologically rich language?**
Since BERTopic relies on an embedding approach which takes context into account, in theory, it should provide informative results even without significantly changing the text structure first. In the case of tweets, researchers in [4] and [7] claim that employing minimal preprocessing is sufficient for English. Proving that BERTopic can overcome the need for more thorough preprocessing (lemmatization) in morphologically rich languages would make topic modeling for Serbian a more straightforward task as it would: 1) assume minimum human involvement in text preparation, 2) prevent relying on restoration of diacritics and lemmatization, which can possibly create faulty lemmas and change the structure of topics.

**RQ 2: How does BERTopic compare to LDA and NMF on lemmatized text?**
Although [2] found some differences in topic quality and stability when using LDA and NMF, their performance was quite similar. We wanted to explore if, under the same conditions, BERTopic would give more informative topics in terms of keywords, and offer some new insights.

## 3.2 The model

BERTopic has a modular architecture made up of several core layers that can be built upon. Those layers are sequential and include embedding extraction, dimensionality reduction and clustering, and creation of topic representation. We discuss the details of our architecture (Fig. 1) in the following subsections.

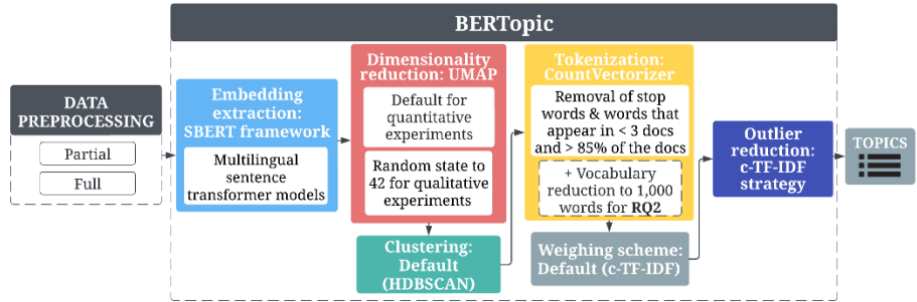

**Fig. 1.** Workflow and details of our BERTopic architecture.

**Embedding models.** Although BERTopic supports the implementation of several different embedding techniques, by default it uses sentence transformers [17]. They are often optimized for semantic similarity, which can significantly help in the clustering task. Since there is no sentence transformer trained solely on Serbian, we could either try to optimize a word embedding model for Serbian or use one of the available multilingual sentence transformers that were trained on multilingual data, including Serbian. In all our experiments, we used a sentence transformer architecture. The three multilingual sentence transformer models that are trained on parallel data for 50+ languages including Serbian are:

- *distiluse-base-multilingual-cased-v2*: knowledge distilled version of multilingual Universal Sentence Encoder that encodes the sentences into 512-dimensional dense vector space. Its size is 480 MB (135 million parameters).
- *paraphrase-multilingual-MiniLM-L12-v2*: multilingual version of paraphrase-MiniLM-L12-v2 that maps sentences to a 384-dimensional dense vector space. The size of this model is 420 MB (117 million parameters).
- *paraphrase-multilingual-mpnet-base-v2*: the largest model with the size of 970 MB (278 million parameters). It is a multilingual version of paraphrase-mpnet-base-v2 that maps sentences to a 768-dimensional dense vector space.

**Dimensionality reduction and clustering.** To reduce the embedding dimensionality, we used UMAP, a default BERTopic dimensionality reduction algorithm. We formed the clusters using HDBSCAN with default parameters.

**Creation of topic representation.** CountVectorizer, as the default BERTopic vectorizer model, and c-TF-IDF, which models the importance of words in clusters instead of individual documents, are together responsible for extracting topic representations from the previously created clusters of documents. We used

CountVectorizer to define word filtering options: stop words, filtering of the words that appear in less than 3 tweets and in more than 85% of the tweets.

**Outlier reduction.** When used with HDBSCAN, BERTopic creates a bin for topic outliers, which can sometimes contain over 74% of the dataset [6]. To prevent this, the outlier reduction step can optionally be added on top of the BERTopic architecture. We reduced the outliers to almost 0 using the *reduce_outliers* BERTopic function with c-TF-IDF as the reduction strategy.

### 3.3 The data

To fit the model, we used the dataset created for the study presented in [2]. The dataset is composed of 3,286 tweets that express negative attitudes towards COVID-19 vaccination in Serbia. The authors report that the dataset was manually checked for topics, therefore we knew that it contains 15 broad topics. We applied two preprocessing scenarios to the tweets:

1. Partial preprocessing, which consisted of transliterating from Cyrillic to Latin, removal of links, mentions and emojis, conversion of hashtags into words, removal of numbers and punctuation, and lowercasing.
2. Full preprocessing, which consisted of partial preprocessing and lemmatization.

Transliteration was performed using *srtools* [18], lemmatization was performed using the *classla* library for non-standard Serbian [19], and we used custom Python regex for the remaining preprocessing steps.

### 3.4 Evaluation

We performed quantitative evaluation of our models using two metrics – topic coherence (TC) and topic diversity (TD) – both commonly used to evaluate topic models [4, 6, 20]. According to [21], TC represents average semantic relatedness between topic words. The specific flavor of TC we used was NPMI [22]. NPMI ranges from -1 to 1, where a higher score signifies that words in a topic are more strongly related. TD [20] measures the percentage of unique words in the top-n words across the topics. It ranges from 0 to 1, where 1 signifies more varied and 0 indicates redundant topics.

For RQ1, we evaluated all three BERTopic embedding models in two preprocessing settings – partial preprocessing and full preprocessing, as defined in Section 3.3. For RQ2, we compared TC and TD between BERTopic models with LDA and NMF. Since both LDA and NMF require fully preprocessed text, we used that preprocessing scenario for all the models. Researchers in [2] compare LDA and NMF using the vocabulary reduction to 1,000 words, so we set the same parameter in BERtopic models (see **Fig. 1**) for the sake of comparability. In both experiments, we averaged TC and TD across 3 runs for 10-50 topics with steps of 10 for every model.

Besides using TC and TD as quantitative measures, we also manually checked the topics for keyword diversity, overall interpretability, and novelty. To obtain repeatable results, we set the UMAP random state to 42. For both research questions, we compare the topics and keywords of best performing models after quantitative evaluation. By default, BERTopic does not put any limitations on the number of topics, which can result in hundreds of topics for larger datasets [7]. We predefined this number in both RQ for the sake of comparability. For RQ1, we set the number of topics to 15 for both models to match the number of topics manually identified by [2]. For RQ2, we matched the most optimal number of topics of the best performing traditional topic model. We also reduced the vocabulary to 1,000 words for the same reason.

## 4  Results & Discussion

### 4.1  RQ1

**Quantitative evaluation**. For partially preprocessed text, the third and the largest model gave the best TC (-.133) and TD (.896) scores (see Table 1). The other two embedding models share the TC score of -.145. The second-best TD score was achieved by *paraphrase-multilingual-MiniLM-L12-v2*. For fully preprocessed text, *distiluse-base-multilingual-cased-v2* had the best results for TC and it shares the best TD score with our second model. All three models achieve high TD values in both preprocessing scenarios, suggesting diverse keywords regardless of the model. TD scores are slightly higher for partially preprocessed text, with *paraphrase-multilingual-mpnet-base-v2* achieving the highest one. On the other hand, TC scores are slightly better for all three models in the case of fully preprocessed text, suggesting more coherent topics.

**Table 1.** Topic coherence (TC) and topic diversity (TD) for different BERTopic embedding models and two preprocessing scenarios.

| BERTopic embedding model | Partial preprocessing | | Full preprocessing | |
|---|---|---|---|---|
| | TC | TD | TC | TD |
| distiluse-base-multilingual-cased-v2 | -.145 | .887 | **-.042** | **.868** |
| paraphrase-multilingual-MiniLM-L12-v2 | -.145 | .895 | -.063 | **.868** |
| paraphrase-multilingual-mpnet-base-v2 | **-.133** | **.896** | -.058 | .860 |

**Qualitative evaluation.** We compared the models with the highest TC and TD scores per preprocessing scenario during the quantitative evaluation: *paraphrase-multilingual-mpnet-base-v2* for partially preprocessed text and *distiluse-base-multilingual-cased-v2* for fully preprocessed text. To start, we paired together the topics yielded by the two models based on keywords. In Table 2 we show five illustrative topics that cover over 60% of documents in each scenario. By looking at 10 representative keywords, we can see that even without lemmatization we obtained informative topics.

As for keyword diversity and interpretability, except for several morphological variations of three nouns and a verb in the keywords of the first model (underlined), the words are varied for both models. The bolded keywords are the ones that clearly point to the interpretation of each topic. For example, the fourth topic shows less keyword diversity for the partially preprocessed text since there are four different morphological forms of the same word (*dete* – child) in the top 10 keywords. Despite this, the keywords are still informative and indicate the concern over the side effects of mandatory vaccination for children. While keywords in this topic for fully processed text may seem more varied, the number of tweets and inspection of representative documents prove that this variety stems from several distinct topics merged into one. The same can be noticed in the first topic, but this time for partially preprocessed text.

Judging by the results, it seems that parameters need to be separately defined for different levels of preprocessing. Applying the same parameters to both scenarios affects the keywords, which is reflected in the number of documents as well. However, even under these conditions, the keywords are diverse and informative for both preprocessing scenarios, indicating that BERTopic can be successfully applied to partially preprocessed text in Serbian.

**Table 2.** Overview of ten keywords and the number of tweets per topic for five illustrative topics obtained by *paraphrase-multilingual-mpnet-base-v2* and *distiluse-base-multilingual-cased-v2*.

| *paraphrase-multilingual-mpnet-base-v2* | | *distiluse-base-multilingual-cased-v2* | |
|---|---|---|---|
| Keywords | No. of tweets | Keywords | No. of tweets |
| **eksperiment** (experiment), **nuspojave** (side effects), zna (knows), sto (hundred), **dnk** (DNA), **mrna** (mRNA), godina (year), dr (dr), ce (will), **bil** (Bill) | 1097 | **eksperimentalan** (experimental), **ispitivanje** (examining), faza (phase), lek (drug), medicinski (medical), nijedan (none), proći (pass), struka (experts), **nuspojava** (side effect), **proizvođač** (producer) | 215 |
| **imunitet** (immunity), **zaštita** (protection), **simptomi** (symptoms), **koronu** (corona), imam (I have), **štiti** (protects), **korone** (corona), **vakcinaciju** (vaccination), **antitela** (antibodies), **prirodni** (natural) | 451 | **virus** (virus), **imunitet** (immunity), **soj** (strain), **simptom** (symptom), grip (flu), **štititi** (protect), napraviti (to make), hiv (HIV), **prirodan** (natural), **preležati** (to develop immunity) | 606 |
| **smrti** (death), **slučajeva** (cases), **umrli** (died), godine (year), **umrlo** (died), broj (number), **smrtnih** (death, adj), **smrt** (death), miliona (million), **umrlih** (dead) | 212 | **umreti** (to die), broj (number), **bolnica** (hospital), **umirati** (die), slučaj (case), **ubiti** (to kill), **smrtan** (deadly), **zaraziti** (to infect), **respirator** (respirator), tri (three) | 263 |
| **decu** (children), **deca** (children), **dece** (children), **štete** (damage), **pravo** (right), **deci** (children), vakciniše (vaccinates), svoju (one's own), vakcinacije (vaccination), vakcinom (vaccine) | 153 | **dete** (child), **gejts** (Gates), **bil** (Bill), misliti (think), **dnk** (DNA), **nuspojava** (side effect), srbija (Serbia), **posledica** (consequence), narod (people), **menjati** (change) | 1035 |
| **maske** (masks), **štite** (protect), **maska** (mask), nose (wear), **distanca** (distance), **mere** (measures), nosi (wears), **štete** (damage), ratom (war), svjet (world) | 94 | **maska** (mask), nositi (to wear), **distanca** (distance), **štititi** (to protect), dobro (well), **mera** (measure), **odbiti** (to reject), **pasoš** | 144 |

| | | (passport), naravno (of course), ruka (hand) | |
|---|---|---|---|

## 4.2 RQ2

**Quantitative evaluation.** The results in Table 3 show that differences in TD and TC scores between BERTopic models are slight, with *distiluse-base-multilingual-cased-v2* performing best, as in the case of RQ1. Compared to LDA and NMF, BERTopic achieved better TC scores. NMF performed slightly worse (-.065 compared to the lowest BERTopic score of -.054), while LDA showed a more significant drop (-.104). Even though our TD scores were high for all the three BERTopic models for Serbian dataset, LDA achieved a slightly better result (.897). The only model with a TD score lower than .8 was NMF. Comparing RQ2 results to the TC and TD scores for the RQ1 (Table 1), it can be concluded that vocabulary reduction only slightly influenced these metrics in case of fully preprocessed text.

**Table 3.** Comparison of TC and TD values across three BERTopic embedding models, LDA and NMF on fully preprocessed text and with vocabulary reduced to 1,000 words.

| Model | TC | TD |
|---|---|---|
| distiluse-base-multilingual-cased-v2 | **-.050** | .861 |
| paraphrase-multilingual-MiniLM-L12-v2 | -.054 | .859 |
| paraphrase-multilingual-mpnet-base-v2 | -.051 | .858 |
| LDA | -.104 | **.897** |
| NMF | -.065 | .795 |

**Qualitative evaluation.** We extracted the topics generated by the best performing BERTopic model during RQ2 quantitative evaluation, which is *distiluse-base-multilingual-cased-v2*. We set the number of topics for BERTopic to 14 to match the most optimal number of topics for LDA for this dataset [2], since LDA showed the best TD score during quantitative evaluation.

Authors in [2] gather all the LDA and NMF topics into five groups and 16 subgroups that we used to name and align BERTopic topics (Table 4). When defining the topic names, [2] looked at 20 keywords and representative documents, which we did as well. Topics detected by BERTopic match the ones in [2] in 69% of cases, meaning that 31% of the topics found by LDA and NMF were not detected by BERTopic. BERTopic did not isolate any topics that deal with the number of doses, which were the topics detected by both LDA and NMF. One BERTopic topic contains several topics in one, combining a conspiracy theory of DNA change with concern over not having a choice with vaccinating children and doubt about effectiveness for new strains, similarly as in RQ1. In the case of the topic dealing with mistrust of authorities, BERTopic breaks it into a total of eight topics, each covering a specific aspect of mistrust. One of them is a completely novel topic that groups tweets regarding concerns about different vaccine manufacturers (84 tweets). While some topics seem uninterpretable by looking at the keywords (e.g., third topic under *Mistrust of government and political decision makers*), there is a clear idea of the topic based on the keywords for most topics.

With the same parameter settings, but without limiting the number of topics, BERTopic found 41 topics for this dataset. By closer inspection of these topic representations, BERTopic detected all reasons identified by [2] as separate topics. Although this number of topics is not appropriate for this dataset, this approach could be an important starting point for further analysis and parameter optimization, as it could provide more detailed or new insights into topics hidden in the dataset. In some cases, identifying smaller topics may be very important in some fields.

Another important BERTopic parameter that significantly affects the number of generated topics is the minimum topic size (*min_topic_size*), which is the minimum number of documents to form a topic. The default parameter value in BERTopic, which we used in our experiment, is 10. Increasing this value results in a lower number of clusters/topics when HDBSCAN is used as the clustering algorithm. When we set this parameter to 15, BERTopic generated 23 topics without any outliers.

To answer RQ2, BERTopic's performance is comparable to LDA and NMF on this dataset, but not under the same parameter settings. When the number of topics is not limited and when the minimum topic size parameter value is changed, BERTopic could potentially provide new insights.

**Table 4.** LDA and NMF topics for the COVID-19 dataset (taken from [2]) and the corresponding BERTopic topics with the number of tweets.

| Reasons for vaccine hesitancy identified by LDA and NMF | BERTopic topic representations | No. of tweets |
|---|---|---|
| Concern over general side effects | umreti (to die), smrt (death), nuspojava (side effect), umirati (to be dying), bolnica (hospital), slučaj (case), život (life), smrtan (deadly), tri (three), nevakcinisan (unvaccinated) | 427 |
| Concern over side effects for children | dete (child), gejts (Gates), bil (Bill), nov (nov), dnk (DNA), soj (strain), bolest (illness), menjati (change), priča (story), sloboda (freedom) | 962 |
| Concern over side effects due to many required doses | | - |
| Concern over vaccine effectiveness: natural immunity is better protection | virus (virus), imunitet (immunity), simptom (symptom), hiv (HIV), otrov (poison), prirodan (natural), napraviti (make), štititi (protect), zaraziti (infect), bolest (illness) | 517 |
| Concern over vaccine effectiveness: vaccines are not effective against new COVID-19 strains | dete (child), gejts (Gates), bil (Bill), nov (nov), dnk (DNA), soj (strain), bolest (illness), menjati (change), priča (story), sloboda (freedom) | 962 |
| Concern over vaccine effectiveness: vaccines are not effective since so many doses are required | | - |
| Concern over side effects of insufficiently tested vaccines | | - |
| Concern over effectiveness of insufficiently tested vaccines | eksperiment (experiment), eksperimentalan (experimental), ispitivanje (examining), faza (phase), testiranje (testing), vršiti (perform), medicinski (medical), trajati (last), pravo (right), nijedan (none) | 241 |

| Violation of freedom by imposing the use of insufficiently tested vaccines | dete (child), gejts (Gates), bil (Bill), nov (nov), dnk (DNA), soj (strain), bolest (illness), menjati (change), priča (story), sloboda (freedom) | 962 |
|---|---|---|
| Mistrust of medical experts and institutions | nauka (science), bog (god), naučan (scientific), vera (faith), dokazati (prove), laž (lie), dokaz (proof), verovanje (belief), reč (word), govoriti (speak) | 209 |
| | dr (dr), doktor (doctor), lekar (doctor), medicina (medicine), medicinski (medical), antivakser (anti-vaxxer), nauka (science), mrn (mRN), efikasnost (effectiveness), crn (black) | 240 |
| | maska (mask), nositi (wear), distanca (distance), štititi (protect), odbiti (refuse), mera (measure), značiti (to mean), naravno (of course), virus (virus), daleko (far) | 138 |
| | Mesec (month), javan (public), zdrav (healthy), zakon (law), radnik (worker), javno (public), vlada (government), pitati (ask), doneti (bring), zdravstven (health, adj) | 44 |
| Mistrust of government and political decision makers | kineski (Chinese, adj), ruski (Russian, adj), kinez (Chinese), eksperiment (experiment), sinopharm (Sinopharm), brat (brother), astrazeneca (AstraZeneca), član (member), predsednik (president), testirati (test) | 84 |
| | milion (million), epidemija (epidemics), milijarda (billion), svinjski (swine), država (state), režim (regime), suzbiti (supress), kupiti (buy), hiljada (thousand), isplatiti (pay off) | 101 |
| | fašist (fascist), otrovan (poisonous), kreten (jerk), globus (globus), tv (TV), fašizam (fascism), rnk (RNA), kolonija (colony), zombirati (to zombie), glup (stupid) | 95 |
| | pasoš (passpost), ukinuti (to cancel), smisao (meaning), rio (Rio [Tinto]), obavezan (obligatory), mera (measure), ostati (to stay), glupost (stupidity), nuditi (to offer), zdravlje (health) | 54 |
| Vaccines are a money-making scheme | - | - |
| Vaccines, especially mRNA vaccines, change DNA | dete (child), gejts (Gates), bil (Bill), nov (nov), dnk (DNA), soj (strain), bolest (illness), menjati (change), priča (story), sloboda (freedom) | 962 |
| COVID-19 does not exist; thus, vaccines are unnecessary | - | - |
| Vaccines are a means of population reduction and control | čekati (wait), nadati (to hope), panika (panick), očekivati (expect), nov (new), red (line), zaraditi (earn), mera (measure), depopulacija (depopulation), strah (fear) | 64 |
| Vaccines are an instrument of world powers and their agenda | nemački (German), rat (war), rus (Russian), ukrajina (Ukraine), ruski (Russian), rusija (Russia), promoter (promoter), agenda (agenda), idiot (idiot), shvatiti (realize) | 106 |

## 5  Conclusions & Future work

In this paper, we tested the performance of BERTopic on short text in Serbian. We were interested in whether BERTopic can yield meaningful topics when applied to morphologically rich slightly processed short text and how well it performs in comparison with LDA and NMF on fully processed text. To answer the first question, we compared the performance of BERTopic with different embedding models on fully and partially preprocessed text and found that BERTopic can produce meaningful and informative topics even with slight preprocessing. In this case, the larger model, the better the performance. As for the second question, we concluded that applying the same parameters as the ones used for LDA is not the optimal scenario for BERTopic. When the number of topics was not limited, BERTopic was able to provide novel insights.

There are several directions we would like to explore in the future. We plan to apply BERTopic to different datasets to check if our conclusions can be generalized. We also plan to explore its prediction capabilities on new, unseen documents. Since BERTopic supports using different transformer models as the embedding models, we also plan to test the applicability and performance of the currently only language model trained on Serbian data – BERTić [23].